# Ladder: A software to label images, detect objects and deploy models recurrently for object detection

Zhou Tang[1*], and Zhiwu Zhang[1*]

[1]Department of Crop and Soil Sciences, Washington State University, Pullman, Washington, 99164 USA.
*Corresponding authors: Zhou.Tang@WSU.edu and Zhiwu.Zhang@WSU.edu

**Abstract**

Object Detection (OD) is a computer vision technology that can locate and classify objects in images and videos, which has the potential to significantly improve efficiency in precision agriculture. To simplify OD application process, we developed Ladder – a software that provides users with a friendly graphic user interface (GUI) that allows for efficient labelling of training datasets, training OD models, and deploying the trained model. Ladder was designed with an interactive recurrent framework that leverages predictions from a pre-trained OD model as the initial image labeling. After adding human labels, the newly labeled images can be added into the training data to retrain the OD model. With the same GUI, users can also deploy well-trained OD models by loading the model weight file to detect new images. We used Ladder to develop a deep learning model to access wheat stripe rust in RGB (red, green, blue) images taken by an Unmanned Aerial Vehicle (UAV). Ladder employs OD to directly evaluate different severity levels of wheat stripe rust in field images, eliminating the need for photo stitching process for UAVs-based images. The accuracy for low, medium and high severity scores were 72%, 50% and 80%, respectively. This case demonstrates how Ladder empowers OD in precision agriculture and crop breeding.

## Introduction

The implementation of computer vision technology has significantly enhanced the efficiency of precision agriculture. Object Detection (OD) is one such technology that enables the identification of objects and their locations in images and videos (Redmon et al., 2015). OD has been effectively utilized in various aspects of precision agriculture, such as fruit detection (Koirala et al., 2019; Liu et al., 2020), pet detection (Hong et al., 2021; Rustia et al., 2020; Wang et al., 2021), vegetable detection (Keller et al., 2018), and crops yield detection (Gong et al., 2020; He et al., 2020; Li et al., 2021). Among the algorithms used, You-Only-Look-Once (YOLO) (Redmon et al., 2015) is highly preferred because it can make accurate predictions in a single network with one evaluation, allowing for high detection speed in real-time detection videos. However, preparing a training dataset for YOLO can prove to be a time-consuming process. Researchers who obtain a trained YOLO also require a convenient GUI for deploying their model for future use. As new images are being generated continuously under varying conditions, re-training of the YOLO model can prove to be challenging for researchers.
The OD have been applied to the Unmanned Aerial Vehicles (UAVs) based images in many applications (Ramachandran and Sangaiah, 2021). Though UAVs offer an efficient and flexible way to capture images, it still brings challenges to OD, such as changing of angle, resolution and illumination. In the pipeline of analyzing UAV-based image data in agriculture, stitching images is almost a standard step (Tsouros et al., 2019). However, this process typically reduces image resolution and can causes unpredictable distortion in final image product, thereby limiting the performance of subsequent computer vision models. Extracting pixels of interesting manually is another time-communing process after stitching (Meivel and Maheswari, 2021). While some software provides a semi-automatic to reduce the pain of manually drawing area of interesting (Tresch et al., 2019), it may also reduce image resolution. Even after creating a final classification model through these steps, it can still be challenging to use for future users. They need to learn all software in the analyzing pipeline, from preprocess images to obtaining classification results.

To address these challenges, we present an open-source software called Ladder which aims to reduce barrier to implementing OD and UAV-based RGB image. Ladder provides a user-friendly graphic user interface (GUI) within the popular Python environment, making it easy for researchers to create training datasets, train YOLO models, and deploy them seamlessly in production environments. Ultimately, Ladder can help advance more fields, such as precision agriculture, by enabling researchers to leverage the latest computer vision technologies effectively.

**Methods**

*Implementation*

Ladder is written with Python 3.6 which can run on Mac OS, Windows and Linux operating systems. It requires a minimum of an 8 GB RAM and 64-bit processor (https://github.com/Mrwow/Ladder). The GUI of Ladder has been derived from the Labelme software (https://github.com/wkentaro/labelme). The YOLOv3 training process and detection pipeline are all based upon the YOLOv3 repository (https://github.com/ultralytics/yolov3). The Ladder GUI consists of three main parts: a top menu bar, a label list on the left, and a canvas on the right (**Fig. 1a**). Users can load an image via the top menu bar and select an appropriate size by zooming in or out.

There are two options for labeling the image. The first option is to manually draw rectangles by selecting "Draw" in the top menu bar. Users can define these rectangles directly by placing two anchor points at the top left and bottom right corners, respectively. In the newly opened dialog window, users can add a label for each rectangle. The second option is to use a pre-trained model for initial predictions. Users can load a weight file for the YOLO model by clicking "Detect" in the top menu bar, which will automatically draw rectangles with corresponding labels on the image. The accuracy of these rectangles depend on how well the YOLO model has been trained with labeled data. Currently, the Ladder only supports YOLO version 3 model. This function enables users to deploy their own YOLO model to detect future images.

Users can edit existing rectangles manually drawn or predicted by YOLO after clicking "Edit" in the top menu bar. Each selected rectangle will be hightlighted with a single click. Users can easily adjust the position and size of the selected ractangle by moving its anchor points. Upon right-clicking the selected rectangle, users can edit its label in a newly opened dialog window. Same category instances' label should remain coincident. When annotation is complete, all rectangles with their labels can be save into a json file for future use. When loading an image into the canvas, a json file with same name storing labeled boxes will be also loaded. Another option for labeled images is to retrain the OD model, which can be easily done by clicking the "Train" in the top menu bar. Users can indicate where the training data is, and a YOLO v3 model will be retrained. The prediction, edition and update model are a recurrent circle (**Fig. 1b**) that gradually improves the capability of the OD model.

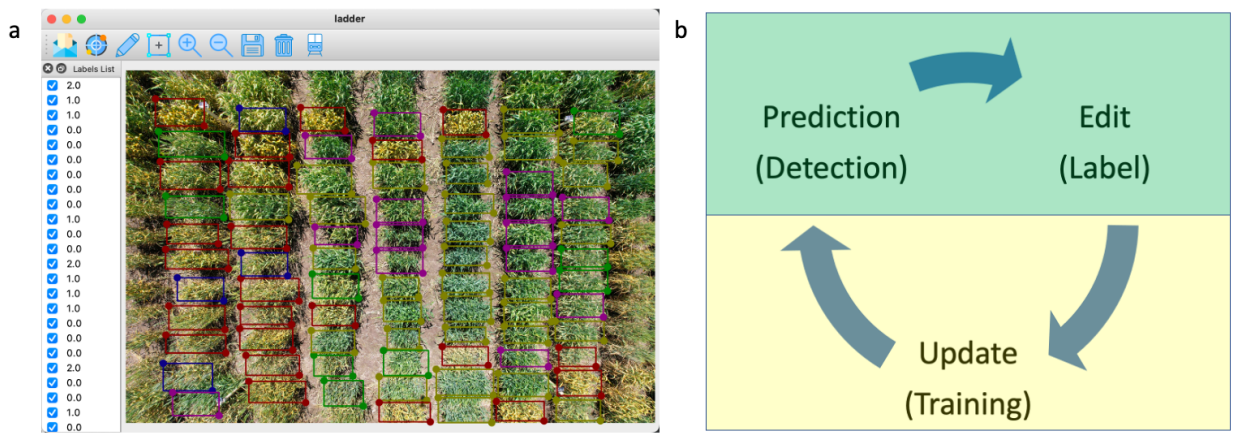

**Figure 1.** Ladder's graphic user interface (GUI) and main functions. The GUI have three parts (a). It can achieve a pipeline of prediction, edit and update for an object detection model development (b).

## Use case

We conducted a survey of a wheat stripe rust nursery field using DJI air2s, flying at a height of around four meters. Total 900 single-row spring wheat plot were planted in the Spillman Agronomy Farm (46°41'41.1 "N, 117°08'30.3 "W) (https://striperust.wsu.edu/disease-management/fungicide-data/). Each plot was assigned a score from 0 to 100 to indicate the severity of wheat stripe rust. The plots were categorized into three groups: low (0-20), moderate (21-60) and high (61-100) groups. A total of 48 RGB images were collected, and they were evenly divided into training and testing sets. We used the Ladder to label the first 24 images and fine tune the YOLO v3 that was pre-trained with the COCO dataset (Lin et al., 2015). The updated YOLOv3 model was then used to assist in labeling the remaining 24 images.

A confusion matrix heatmap was generated to show the accuracy of the trained YOLOv3 in assessing the different levels of wheat stripe rust severity (**Fig. 2a**). Additionally, the precision-recall curve demonstrated the area under the curve (AUC) for each group (**Fig. 2b**). With a class prediction confidence threshold set at 0.25 and a predicted box overlap set at 0.45, the prediction accuracy for the low, moderate, and high groups was 72%, 50% and 80%, respectively (**Fig 2a**). However, some single-row plots were missed, with 21%, 20% and 15% of low, moderate and high group, respectively. Given the limited training data for fine-tuning the YOLOv3 model, it is not surprising that some plots were predicted as background. From the precision-recall curve

(**Fig 2b**), we found current re-trained YOLOv3 showed much higher capability in detecting low and high group than moderate class.

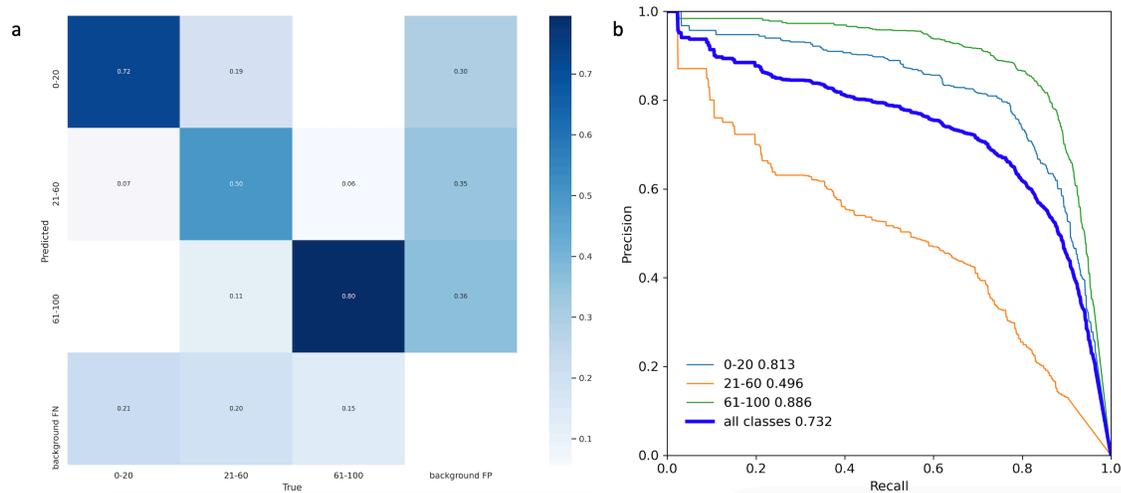

**Figure 2**. Performance of YOLO v3 trained under the Ladder scheme. A heatmap of the confusion matrix is used to display the accuracy of trained YOLO v3 on the assessment of various levels of wheat stripe rust severity score: low (0-20), moderate (21-60), and high (61-100) (a). The precision-recall curve to demonstrate area under the curve (AUC) under assessment of each group (b).

## Conclusions

Ladder provides a user-friendly GUI that allow users to annotate images, detect objects and retrain OD model, which can significantly reduce the barriers of applying OD. Researcher would also get higher resolution training images from UAV-based camera by skipping images stitching if they include the Ladder in the pipeline of preprocessing images.

## Author Contributions

Conceptualization, Z.Z., Z.T.; Software, Z.T.; Formal analysis, Z.T.; Writing-original draft, Z.T.; Writing-review and editing, Z.Z.; supervision, Z.Z.; Project administration, Z.Z.; Funding acquisition, Z.Z. All authors have read and agreed to the published version of the manuscript.

## Conflicts of Interest

The authors declare no conflict of interest.